# MOANOFS: Multi-Objective Automated Negotiation based Online Feature Selection System for Big Data Classification

Fatma BenSaid, *Member, IEEE*, and Adel M. Alimi, *Senior Member, IEEE*

**Abstract**— Feature Selection (FS) plays an important role in learning and classification tasks. The object of FS is to select the relevant and non-redundant features. Considering the huge amount number of features in real-world applications, FS methods using batch learning technique can't resolve big data problem especially when data arrive sequentially. In this paper, we propose an online feature selection system which resolves this problem. More specifically, we treat the problem of online supervised feature selection for binary classification as a decision-making problem. A philosophical vision to this problem leads to a hybridization between two important domains: feature selection using online learning technique (OFS) and automated negotiation (AN). The proposed OFS system called MOANOFS (Multi-Objective Automated Negotiation based Online Feature Selection) uses two levels of decision. In the first level, from n learners (or OFS methods), we decide which are the k trustful ones (with high confidence or trust value). These elected k learners will participate in the second level. In this level, we integrate our proposed Multilateral Automated Negotiation based OFS (MANOFS) method to decide finally which is the best solution or which are relevant features. We show that MOANOFS system is applicable to different domains successfully and achieves high accuracy with several real-world applications.

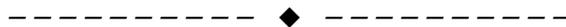

## 1 INTRODUCTION

**D**URING the last three decades, Feature Selection (FS) has been extensively studied in Data Mining [1], [2], Pattern Classification [3], [4] and Machine Learning [5], [6]. FS is defined as the process of selecting a subset of relevant features and removing the redundant ones from a dataset for building effective prediction models.

In recent years, an enormous increase in data (news, medical imaging) has been observed which allows an increase in redundant information.

Even worse, the redundancy of irrelevant data has a negative impact on the performance of classification methods associated. With the rapid development of the Internet, current tremendous amounts of data up to millions or billions, can be collected for training machine learning models.

Most existing studies of feature selection are conducted in batch learning (off-line learning). In the batch learning, all features are given a priori in training instances. Such assumptions may not always hold for some real-world applications. In these real applications, training examples often arrive in a sequential manner, or it is expensive to collect the full information of training data.

With the emerging of large scale data and big data applications, the feature selection based on batch learning methods becomes non-practical.

Recently, Online Feature Selection (OFS) methods [7], [8], [9], [10] have been proposed to face out the drawbacks of batch feature selection methods. In fact, the proposed OFS methods tend to resolve feature selection tasks by exploring online learning techniques in machine learning.

Nowadays, significant parts of the information are stored in textual databases (or text documents) which are composed of a large set of documents from various sources, such as news, articles, books, digital libraries, e-mail messages and Web pages. It is obviously important to consider that a real-world application has to deal with sequential, massive and high dimensional training data. Online learning has been extensively studied in machine learning and data mining [11], [12], [13]. In a traditional online learning task (e.g. online classification), a learner is trained in a sequential manner to predict the class labels of a sequence of instances.

With the development and penetration of distributed data mining within different disciplines, both feature selection and online learning have emerged to enhance techniques of relevant data selection. These data are mined for several identifications of data mining task in an online fashion for efficient knowledge discovery and collaborative computation.

In this paper, we find a solution to the problem of online feature selection with large-scale and ultra-high dimensional data for classification task using a new vision of data analysis. In fact, when we treated an online feature selection problem, questions raised are: is the online feature selection method used to select relevant features the best method that can enhance the classification performance? Can selecting the features let's say 1,2,3,8… give minimum errors than the selection of 1,3,4,8… features!? Can making a combination between some OFS methods or between intelligent methods improve the performance of classification!? These different questions were behind developing a new idea in this paper which is hybridization between two domains aiming to assure decision making: feature selection in online fashion and automated negotiation (which is described with a philosophical vision).

This paper aims to address large-scale online feature selection problems with big data. To this end, we propose a novel online feature selection system by exploring the recent advances of online machine learning techniques [11], [14], [15], and a conflict resolution technique (Automated Negotiation) [16] for the purpose of enhancing the classification performance of ultra-high dimensional databases.

The remainder of this paper is organized as follows. We present a review of OFS and AN in Section 2. Then, we describe our proposed ANOFS methods and OFS system in Section 3 and Section 4 respectively. Finally, we draw a conclusion of this paper and we present possible future work in Section 5.

## 2 REVIEW OF ONLINE FEATURE SELECTION AND AUTOMATED NEGOTIATION

In this section, we first give an overview of online feature selection methods. Then, we introduce the principle of automated negotiation (AN). Specifically, we present some state-of-the-art works using AN on which we were referred to establish our OFS system.

### 2.1 Online Feature Selection

Online Feature Selection (OFS) aims to solve the feature selection problem in an online fashion by effectively exploring online learning techniques. The key challenge of Online Feature Selection is how to make accurate prediction for an instance using a small number of active features. Let us have an overview of online feature selection methods.

One of the most straightforward approaches to online feature selection is applying the Perceptron algorithm via truncation (**PETrun**) [10]. Specifically, at each step, the classifier firstly predicts the label $\hat{y}_t$ with $w_t$. If $\hat{y}_t$ is correct, then $w_{t+1} = w_t$; otherwise, the classifier will update $w_t$ by Perceptron rule to obtain $\hat{w}_{t+1} = w_t + \eta_t y_t x_t$, which will be further truncated by keeping the largest B absolute values of $\hat{w}_{t+1}$ and setting the rest to zero. The truncated classifier, denoted by $w_t^B$ or $w_{t+1}$, will be used to predict the next observation.

Wang et al. [10] observed that the above method cannot guarantee a small number of mistakes since it fails to ensure small numerical values of truncated elements, thus leading to a nontrivial loss of accuracy. Consequently, the authors [10] proposed a novel first-order online feature selection scheme (FOFS) by exploring online gradient descent with a sparse projection scheme before truncation [17], which guarantees the resulting classifier $w_t$ to be restricted into an l1-ball at each step (see Algorithm 1).

---
**Algorithm 1** FOFS: First-order OFS via Sparse Projection
Input: B, η
Following the similar framework as PETrun (Perceptron algorithm via truncation) but use constant learning rate η

$$\widetilde{w}_{t+1} = (1 - \lambda\eta)w_t + \eta y_t x_t$$

$$\hat{w}_{t+1} = min\left\{1, \frac{\frac{1}{\sqrt{\lambda}}}{\|\widetilde{w}_{t+1}\|_2}\right\}\widetilde{w}_{t+1}, \text{where } \lambda \text{ is a regularization parameter}$$

$$w_{t+1} = Truncate(\hat{w}_{t+1}, B)$$

---

Wu et al. [18], [19] proposed an online feature selection method (SOFS) by exploiting second-order information. SOFS uses confidence-weighted (CW), a state-of-the-art method for second-order online learning, and extend it to tackle online feature selection tasks to deal with large-scale ultra-high dimensional sparse data streams.

Yu et al. [20] developed a Scalable and Accurate OnLine Approach for feature selection (SAOLA). SAOLA can filter out redundant features using a theoretical analysis to derive a low bound on pairwise correlations between features.

Yang et al. [21] proposed a limited-memory and model parameter free online feature selection method namely online substitution (OS) in order to overcome two drawbacks of existing methods which solve an $L_1$ norm minimization problem (minimization of total loss). In fact, these two disadvantages are: 1) the penalty term for $L_1$ norm term is hard to choose; and 2) the memory usage is hard to control or predict. OS essentially aims at solving an $L_0$ norm constraint problem:

$$min\ L(Xw, y) \quad s.t.\ \|w\|_0 \leq s$$

where $X \in \mathbb{R}^{n \times p}$ is a feature matrix of n samples with p features, $y \in \mathbb{R}^n$ is the label vector, and s is the total number of features we want to select.

Perkins et al. [22] presented a flexible OFS approach called Grafting. In this work, which treats the selection of suitable features as an integral part of learning a predictor in a regularized learning framework. To make it suitable for large problems, grafting operates in an incremental iterative fashion, gradually building up a feature set while training a predictor model using gradient descent.

Perkins and Theiler [7] tackled the problem in which features arrive one by one at a time instead of all being available from the start. Online Feature Selection (OFS) assumes that it is not affordable to wait, for any reason, until all features have arrived before learning begins. Therefore, one needs to derive a mapping function *f* from the inputs to the outputs that is as "good as possible" using a subset of features just seen so far.

We have to distinguish between online feature selection and online streaming feature selection. We formulate dynamic features as streaming ones, whereby they are no longer static but flow in one by one, and each new feature is processed upon its arrival.

Zhou et al. [7], [23] proposed a streamwise feature selection method which evaluates each feature once when it is generated using information-investing and α-investing (two adaptive complexity penalty methods) to dynamically adjust the threshold on the error reduction required for adding a new feature.

Wu *et al.* [8] presented the OSFS method (Online Streaming Feature Selection) to online select relevant and non-redundant features. The authors in [9] presented a novel framework for selecting features from streaming ones, which is inspired by feature relevance and feature redundancy. In [9], an efficient Fast-OSFS algorithm is proposed to improve feature selection performance. This framework involves two key components: 1) the utilization of feature relevance to select features on the fly, and 2) the removal of redundant features from the selected candidates which are far, based on feature redundancy.

Keerthika and Priya [24] examined various feature reduction techniques for intrusion detection, where training data arrive in a sequential manner from a real-time application.

Wang *et al.* [25] developed a novel online group feature selection method named OGFS which solves the problem of image analysis. In fact, this problem assumes that features are generated individually but there are group structures in the feature stream. This approach consists of two stages: online intra-group selection and online inter-group selection. In the intra-group selection, a criterion based on spectral analysis is designed to select discriminative features in each group. In the inter-group selection, a linear regression model is used to select an optimal subset. This two-stages procedure continues until there are no more features arriving or some predefined stopping conditions are met.

The striving to reduce the dimensionality of the growing data creates new tough challenges in how selecting both the best OFS method and the best combination of features that helps the improvement of classifier performance as well as having a best pattern recognition in decision making result.

Having studied and tested the efficiency of classifiers' combination [26], we have supposed that the same idea will be great and successful in the dimensionality reduction step. Theoretically, we are in front of different propositions, so, we can say that we are in a conflict situation; mainly a conflict between learners. Conerning literature, to solve the conflict of interest between classifiers, we used the technique of automated negotiation.

## 2.2 Automated Negotiation

Automated Negotiation (AN) is a technique of conflict resolution. In fact, several definitions of negotiation have been proposed in literature [27], [28], [29]. In the context of classification decision making, AN is used to resolve decision conflict between two or more classifiers. Each with its preferences, seeks to reach an agreement that assign a class to an input pattern.

The work presented in [28] shows a generic framework which defines the different steps and components to conduct automated negotiation process. First of all, we have to define the object of negotiation and specify the number of issues. Then, we should define our negotiators, their preferences, their parameters and the type of interaction between them which leads to bilateral negotiation in the cases of having two negotiators or multilateral negotiation if we have more than two participants. Also, we have to clarify the strategy of each negotiator; this means the different actions to do along the negotiation process. To communicate between these participants, we have to specify rules of encounter or the protocol of negotiation. Finally, the negotiation outcome is specified if participants are cooperative or collaborative.

Several reviews on automated negotiation [16], [30], [31], [32] give the reader more information and help better understanding negotiation principle.

An extensive review of all problems in automated negotiation has nonsense since the amount of literature in this domain is vast and immense. Thus, in this part of literature review we have mainly focused on identifying two types of negotiation models: bilateral and multilateral models to solve the classification problem.

**Bilateral Automated Negotiation.** We give an example of two classifiers that fall into decision conflict of assigning a class to an input pattern. Suppose that this input pattern is the handwritten digit '3'. The first classifier C1 assigns '8' to this input and the classifier C2 decides that this pattern is '3'. The two classifiers are in decision conflict so we decide to integrate a negotiation process in the classification system to resolve this conflict. We notice here that the negotiation objective is the classification of a handwritten digit. The number of negotiators is two so we lead a *bilateral negotiation*. Each participant in this negotiation process has to maximize its utility. For any agreement, this utility can be calculated according to the negotiation participant's utility function. A *utility function* describes a negotiator's preferences, which allows *negotiation outcomes* to be evaluated and compared.

In bilateral automated negotiation, maximum utility for a single agent can become minimum utility for opponent agent, and therefore the chance of agreement is low (see Fig. 1).

If we consider that these negotiators are collaborative, it is worthy to say that they have the same objective: maximization of the classification performance. So, we are in the case of *simple issue negotiation* which is the 'classification performance'. If we suppose now that these negotiators aim to 'maximize the classification performance' and to 'maximize the trust', we notice that we have more than one issue or objective and we call this type of negotiation *multi-issue negotiation*. This makes negotiation much more complex especially when one participant has no information about its opponent.

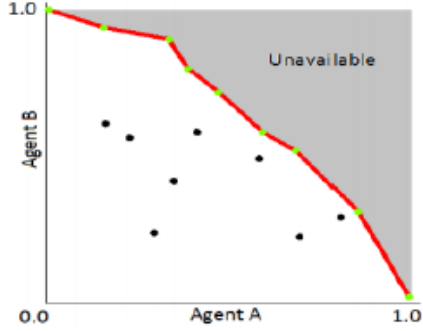

Fig. 1. Utility function in bilateral Negotiation (A point indicates the utility for both agents of a bid. The red line is the Pareto optimal frontier, the green points are the best solutions to each agent).

If we consider $v_1$ and $v_2$ two interdependent issues, then:
$$U_p(\langle v_1, v_2 \rangle) \leq w_{p,1}.U_p(\langle v_1, \emptyset \rangle) + w_{p,1}.U_p(\langle \emptyset, v_2 \rangle) \quad (1)$$
where $w_{p,i}$ is the weight of issue $i$ to negotiator $p$ and $U_p$ is the $p$'s utility function.

In some other negotiations, issues are independent. In this case, the overall utility of the offer is equal to the weighted sum of utilities of different issues. Such utility functions can be defined formally as:
$$U_p(o) = \sum_{i=1}^{n} w_{p,i}.U_{p,i(v_i)} \quad (2)$$
where $o$ is the offer, $n$ is the number of negotiation issues, $w_{p,i}$ is the weight of issue $i$ to negotiation participant $p$ and $U_{p,i}(v_i)$ is the $p$'s utility of issue $i$ for the value of $v_i$. Formally $U_p(o) \in [0,1]$.

To perform such negotiation, negotiators must choose rules to make offers and counter offers. These rules are called negotiation protocol [33]. On the basis of this protocol, each negotiator chooses its strategy [34] which is a specification of the sequence of actions the agent plans to make during the negotiation process.

When the number of participants in a negotiation process exceeds two, we call this type of negotiation Multilateral Negotiation.

**Multilateral Automated Negotiation.** In the most general situation of multilateral negotiation with many issues and with vague preferences, we suppose that $m$ participants take part in the negotiation and the negotiation subject can be characterized by $n$ issues, all of them have numerical nature.

Let $Xij$ denote the value for issue $j$ ($j = 1,...,n$) offered to the negotiation participant $i$ ($i = 1,...,m$) by another participant at some moment. In general, an interval of values is acceptable by each participant, i.e.,
$$a_j^i \leq x_j^i \leq b_j^i, j = 1;...;n, i = 1..m \quad (3)$$

Different values taken from this interval can have different worth for every negotiation participant. The worth of values of negotiation issues is modeled by scoring functions:
$$S_j^i: [a_j^i, b_j^i] \to [0,1], j = 1;...;n, i = 1..m \quad (4)$$

The bigger the value of a scoring function for a certain value of an issue is, the more suitable it is for the negotiation participant. In a real negotiation, different negotiation issues have different importance in relation to every negotiation participant. To model this, we introduce the notion of relative importance that a participant assign to each issue under negotiation. Let $\omega_{ij}$ be the relative importance of issue $j$, $j = 1,...,n$, for the participant $i$, $i =1,...,m$. For convenience, we assume that the normalized relation is valid:
$$\sum_{j=1}^{n} w_j^i = 1, \; i = 1..m \quad (5)$$

Now we suppose that negotiation participant $i$, $i = 1,...,m$, is given an offer. Because the negotiation is characterized by $n$ issues, the offer can be represented by a vector $x = (x_1, ..., x_n)$.

Using the scoring functions and relative importance of issues under negotiation, we can introduce the notion of a general scoring function:
$$S^i(x) = F^i(x, S_1^i(x), ..., S_n^i(x), \omega_1^i, ..., \omega_n^i),$$
$$F^i: R^{3n} \to R, i = 1..m \quad (6)$$

The exact form of scoring functions depends on a concrete situation. In many cases, linear function can be used to model the utility:
$$S^i(x) = \sum_{j=1}^{n} \omega_j^i S_j^i(x), i = 1..m \quad (7)$$

This situation is the simplest one from the mathematical point of view. If all negotiators use the linear scoring functions, it is possible to compute the optimum value of $x$ giving theoretical value for the "best deal". In a real negotiation, however, the final result achieved in the process of negotiation will depend on negotiation strategies, even in the case of linear scoring functions.

If we give the example of the *Time-dependent strategy*:
$$f(t) = f_1 + \left(\frac{t - t_{Init}}{t_{Max} - t_{Init}}\right)^{1/\beta} (f_2 - f_1),$$
$$f: [t_{Init}, t_{Max}] \to [f_1, f_2] t \in [t_{Init}, t_{Max}] \quad (8)$$
where $f_1$ and $f_2$ are user defined value of the negotiation issue. $t_{Init}$ is the initial time of an agent initialized by the system user. $t_{Max}$ is the time max dedicated to an agent (a negotiator) in a negotiation process (the end of negotiation for an agent $a$). $\beta > 0$ is a parameter that determines how the agent changed the value of issue according to the time. Mediator is a coordinator between agents, who handles and manages communication between agents.

Another interesting work on automated negotiation is proposed in [26]. Hamdani *et al.* [26] integrate negotiation between several classifiers to resolve a decision conflict (which is the right class?) between them. Firstly, authors designed an agent classifier capable to interact with its environment and to negotiate with its neighbors. Secondly, Hamdani *et al.* proposed Intelligent Multiple Decision System (I-MDS) which is a multi-agent system where agents are $n$ classifiers. This proposed system was tested on MNIST dataset, and compared to the 'majority voting rule'. It had in most cases better performance of classification.

To explain more the principle of Automated Negotiation, we present below a basic model of negotiation.

**A Simple Negotiation Model.** We suppose that the negotiation space is represented by $Neg = \langle P, A, D, U, T \rangle$ where:
- *P* represents a finite set of participants;
- *A* represents a set of attributes (issues) used in negotiation and understood by all of agents;

- An attribute domain is specified by $D_{a_i}$ where $D_{a_i} \in D$ and $a_i \in A$;
- $U$ denotes the set of utility functions in which the utility of agent $p$ ($p \in P$) is denoted by $U_p^o \in U$;
- Finally, the deadline for every agent $p$ is represented by $t_p^d \in T$.

In this model, it is assumed that information about $P$, $A$, $D$ is exchanged among the negotiation participants.

A *multilateral* negotiation situation can be modeled as many one-to-one *bilateral* negotiations where an agent $p$ maintains a separate negotiation dialog with an initiator (a broker).

In a negotiation round, each agent will make an offer to the initiator. An offer $\vec{o} = \langle d_{a_1}, d_{a_2}, ..., d_{a_n} \rangle$ is a tuple of attribute values (intervals) pertaining to a finite set of attributes $A = \{a_1, a_2, ..., a_n\}$. Each attribute $a_i$ tales its value from the corresponding domain $D_{ai}$.

The valuation function for each agent $p$ for each attribute $a \in A$ is defined by: $U_p^A: A \to [0,1]$. The valuation function of each agent $p$ for each attribute value $d_a \in D_a$ is defined by: $U_p^{D_a}: D_a \to [0,1]$. The valuations of attributes are assumed normalized: $\sum_{a \in A} U_p^A(a) = 1$.

One common way to measure an agent's utility function for an offer $o$ is as follows:

$$U_p^o(o) = \sum_{a \in A}(U_p^A(a) \times U_p^{D_a}(d_a)) \quad (9)$$

Now let's see how the incoming offers are processed in this model. When the offer $o$ is received from an opponent, it is evaluated to see if it satisfies all of its constraints. This is carried out by computing an equivalent offer, $o_\cong$, as an interpretation about the opponent's proposal $o$. The offer $o_\cong$ is equivalent to $o$ if and only if every attribute interval of $o_\cong$ intersects each corresponding attribute interval in $o$. Once it is computed, it decides whether to accept it according to some criteria.

we consider the deadline in terms of time pressure (*time*), which is applied as a coefficient to previous two components. In this work, we use a simplified version of the polynomial time-dependent decision function in [35] to compute and apply this time pressure:

$$time(t) = 1 - \left(min\left(t, t_p^d\right)/t_p^d\right)^{1/\beta} \quad (10)$$

In this function, $t$ denotes the absolute time or the number of negotiation rounds. $t_p^d$ denotes the deadline for agent $p$. This deadline can also be specified either in absolute time or according to the maximum number of negotiation rounds that is allowed for agent p.

Furthermore, $\beta$ specifies the agent's attitude toward concession. This factor is defined by the user before the negotiation starts and may take positive non-zero values.

We reviewed the existing OFS works, then we gave the OFS problem a philosophical vision considering that OFS is a decision-making problem where decision-makers fall in decision conflict. In literature, Automated Negotiation (AN) is a technique of conflict resolution. For this reason, we choose this technique to resolve the problem of OFS. In fact, this section presents a hybridization between two domains: feature selection domain and automated negotiation domain. This novel vision to state-of-the-art methods contributes in the emergence of a novel approach which is Automated Negotiation based Online Feature Selection (ANOFS). In Section 3, we will describe this approach ANOFS and we will show several experimental results to demonstrate its efficiency compared to state-of-the-art OFS approaches.

## 3 AUTOMATED NEGOTIATION BASED ONLINE FEATURE SELECTION

We propose two types of OFS methods which apply the automated negotiation to resolve the problem of feature selection in an online fashion. The first method is named BANOFS (Bilateral ANOFS) and the second type is named MANOFS (Multilateral ANOFS). We began our idea by two negotiators '*neg=2*' (BANOFS) then we expand our research to treat more complex domains and so we increase the number of negotiators to *neg>2* to get the MANOFS method. In this section, we present the OFS problem. Then, we describe the two types of ANOFS, and we report the experimental results comparing our algorithms to other OFS algorithms.

### 3.1 BANOFS: Bilateral Automated Negotiation based OFS Method

*BANOFS: Description*

Considering that two learners have a decision conflict: which are the appropriate features to be selected in order to enhance the classifier performance? Suppose that *RAND* (see Algorithm 2) and $PE_{trun}$ (see Algorithm 3) are the two participants in the negotiation process that try to select best features.

We consider the error rate or the number of mistakes as the utility function of each negotiator, so the object of negotiation is 'minimizing the error rate'. We suppose that we treat a simple-issue negotiation process.

The idea is integrating the principle of automated negotiation between online feature selection algorithms in order to enhance the classification performance. Many existent works [36], [37] use negotiation between classifiers. In this work, we use negotiation approach between learners. We consider a negotiation domain $x$ that contains d features $\{F = f_1, f_2, ..., f_d \mid d$: dimension of the input vector $x\}$ resulting an outcome space $x'$ which represents all the possible outcomes achievable. The formal definition of a negotiation outcome $O$, also referred to the different offers, can be seen as $O = \{o_1, ..., o_p\}$ where $p$ is the number of participants in the negotiation process and $o_1$ for example is represented as $o_1 = \{w_1, err\_count\}$ means a negotiator offer contains the weight vector (the vector which contains selected features) and the error-rate of prediction.

Assume that we adopt two negotiators (learners) that are completely cooperative, each participant $p$ has complete information about the opponent preference (its utility function).

Each negotiator defines its preference (preference profile). For each issue in the domain, a weight is assigned indicating how important that particular issue is to the negotiator (issue weight).

**Algorithm 2   RAND Algorithm**
**Input**
$B$: the number of selected features
**Initialization**
$w_1 = 0$
**for** $t = 1,2,…,T$ **do**
  Receive $x_t$
  Make prediction $sgn(x_t^\top w_t)$
  Receive $y_t$
  **if** $y_t x_t^\top w_t \leq 0$ **then**
    $\hat{w}_{t+1} = w_t + y_t x_t$
    $v = 0$
    $perm_t = randperm(size(\hat{w}_{t+1}, 1))$
    $c_t = perm_t(1 : B)$
    $v(c_t) = 1$
    $w = \hat{w}_{t+1}\ v$
  **end if**
**end for**

**Algorithm 3   Modified Perceptron PEtrun**
**Input**
$B$: the number of selected features
**Initialization**
$w_1 = 0$
**for** $t = 1,2,…,T$ **do**
  Receive $x_t$
  Make prediction $sgn(x_t^\top w_t)$
  Receive $y_t$
  **if** $y_t x_t^\top w_t \leq 0$ **then**
    $\hat{w}_{t+1} = w_t + y_t x_t$
    $w_{t+1} = Truncate(\hat{w}_{t+1}, B)$
  **else**
    $w_{t+1} = w_t$
  **end if**
**end for**

To communicate between these negotiators, we implement ANOFS protocol.

**ANOFS Protocol**

We have chosen *Contract Net Protocol* [33] as basis to design the protocol used in our OFS System. Fig. 2 shows the ANOFS protocol adapted in our work.

The Scenario of this protocol is: a call for proposal CFP (claim: which participant is ready to participate to solve the OFS problem?) will be sent from the initiator to participants. The participant $p$ sends its proposal $w$ (the weight vector) and its prediction error (err-count). The accepted proposals (vectors with selected features) will be collected and the initiator creates a vector that contains all selected features. This new vector $w_{ANOFS}$ (or $w_{initiator}$) will be send to each participant.

Fig. 3 describes a real case of the proposed BANOFS method. The process begins by a simple $w$ sending; it means each participant sends its weight vector which continues the selected features. The first participant *RAND* sends $w_{RAND}$ and its prediction error $err\_count_{RAND}$ and the second $PE_{trun}$ sends $wPE_{trun}$ and $err\_count_{PE}$ to the initiator.

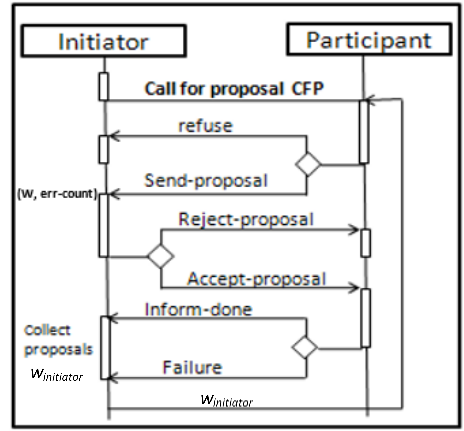

Fig. 2. ANOFS protocol

The initiator follows a set of actions to get its weight vector $w_{ANOFS}$. Suppose that $W_1$ and $W_2$ are the two bids (vector of weights) offered by the two learners (negotiators) $L_1$ and $L_2$.

If $L_1$ chooses a feature that $L_2$ consider irrelevant, the initiator gives chance to this feature and put its weight into $w_{ANOFS}$ and respectively $L_2$ and $L_1$.

If the same feature is selected by the two learners, the initiator has to detect the learner which has minimum error and put its weight into $w_{ANOFS}$. In other words, the initiator accepts the two propositions and sends $w_{ANOFS}$ to each participant. In the second trial ($t = 2$), each participant has to use the $w_{ANOFS}$ (not his saved $w$) to do the prediction step and so on. This idea means to give more chance to each feature selected by the union of all features stayed in *RAND* and $PE_{trun}$ vectors; this increases the confidence rate in each participant and give an opportunity to enhance the classification rate.

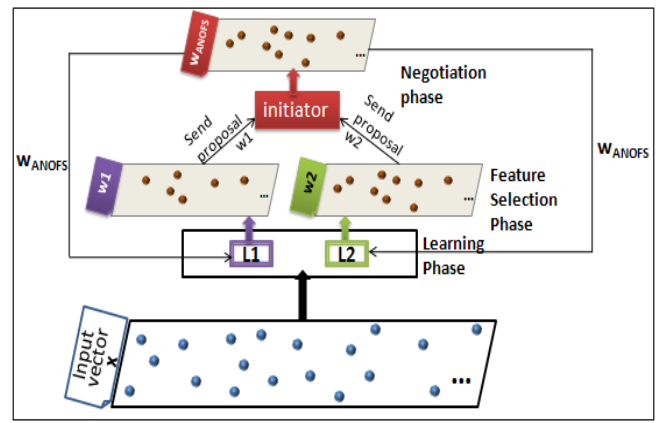

Fig. 3. The proposed Bilateral ANOFS method

The diagram of our negotiation environment is presented in Fig. 4. The ANOFS protocol contains the rules of encounter between the negotiators and the initiator. The actions in the negotiation process are: sending proposal *(w, err_count)*, accepting proposal, refusing proposal. Of course, these actions are followed by some criteria as when the initiator should accept or refuse proposals and when the negotiation should be stopped ($t = tmax$; where $tmax$ is the maximum number of iterations fixed as input).

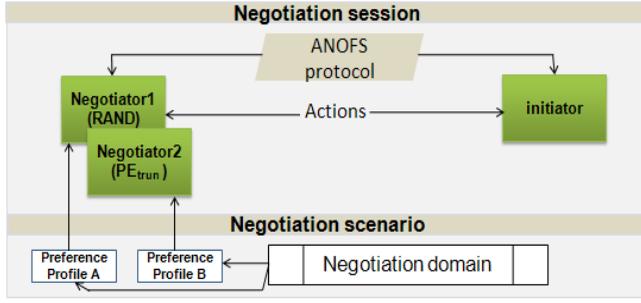

Fig. 4. A diagram of the negotiation environment

We have therefore raised the question: if we have more than two negotiators with several techniques of online learning, can we get better results means can we enhance more the classification performance? This question makes us thinking about a novel method of negotiation that ensures the integration of more than two learners in our negotiation model and this leads to MANOFS method: Multilateral Automated Negotiation based OFS method. For this purpose, we consider online learning algorithms in LIBOL framework [38]. Next, we will describe the MANOFS method.

### 3.2 MANOFS: Multilateral Automated Negotiation based OFS Method

*MANOFS: Description*
We define below our negotiation process MANOFS.
• *Negotiation Object:* what are the relevant features?
*Issues:* we suppose that we have a simple issue which is maximizing the online predictive performance that generates a maximization of classification performance. Of course, we can consider more than one issue in our MANOFS model and in this case, we will speak about multi-issue negotiation process.
• *Which agents participate:* $n$ agent-learners or $n$ OFS methods; a framework of OFS methods.
*Preferences:* all these agent-learner tend to maximize the online predictive performance.
• *Parameters:* each agent has its special parameters and all learner-participants have Trust parameter that indicate the degree of confidence of each learner. Also, the history of each prediction is saved to help the learner in the next turn.
• *Type of interaction:* one-to-many which leads to multilateral negotiation: many bilateral negotiations.
• *Negotiation protocol:* the used protocol is an extended version of the ANOFS protocol presented in [39].
• *Negotiator's strategy:* each leaner accords a weight to relevant feature and 0 to redundant ones. The initiator follows the following steps to create the $w_{MANOFS}$ vector (which is the output of MANOFS):
- If the feature $F_t$ is considered irrelevant by the three learners, $w_{MANOFSt}$ will be equal to 0;
- If the feature $F_t$ is selected by just one between the three learners, $w_{MANOFSt}$ will have the weight given to this feature;
- If more than one learner chooses the feature $F_t$, $w_{MANOFSt}$ will take the weight given by the learner which has the minimum error of prediction.
• *Negotiation outcome:* our agents are cooperative since they opt to find the best solution of feature selection by cooperating all learners.

Fig. 5 presents a real case of the proposed MANOFS method. We describe first of all negotiators: let's consider three participants to the negotiation process where the object of negotiation is the same as cited before; selecting the best relevant features. A vector containing $n$ features is the input to our MANOFS method. $L1$, $L2$ and $L3$ are the three learners that accept the call for proposal of the initiator. At $t=1$ (the first trial), each participant sends its proposal as a weight vector $w$ to the initiator. This vector should contain just the value of the selected features; all the non-selected ones should have zeros as value. Now, the real process of negotiation begins when the initiator considers that there is a difference between different proposals so that we have to negotiate to solve this decision conflict. The initiator follows different actions to do during the negotiation process to get the $w_{MANOFS}$ vector of weights. This vector will be sent to all negotiators. These participants use it in the second trial as source of historic information. Since they will be more trustful, they will even have minimum error of prediction. They can therefore enhance the classification performance. The negotiation process is aborted when the $t=Tmax$.

The question raised during the implementation of this method is: does a feature selected by just one OFS method have the same importance of a feature selected by two or three methods? This question leads us to think about giving more importance to the most selected features; which means according points of trust to each selected feature. In other words, $TF$ (Trust of Feature) is the trust parameter initialized to 0.05 where $TF \in [0, 1]$.

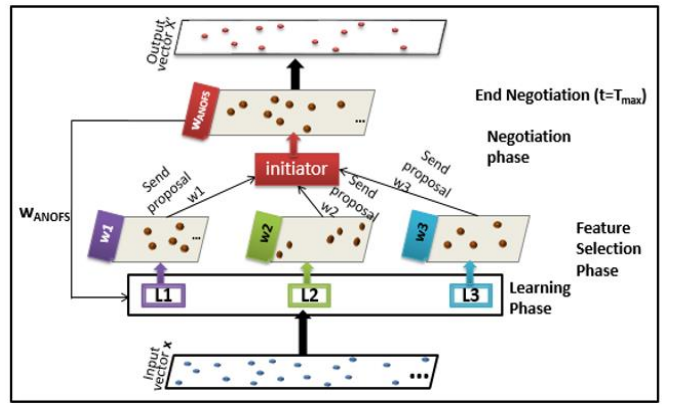

Fig. 5. The proposed Multilateral ANOFS method

If a feature $F_i$ is selected by one method, its $TF$ will be changed as in Equation (11) (see Fig. 6):
$$TF_i = TF_i + \varepsilon \qquad (11)$$
where $\varepsilon$ is another parameter initialized by the system's user. We suppose first of all that $\varepsilon = \frac{1}{n}$ if one participant chooses the feature $i$.

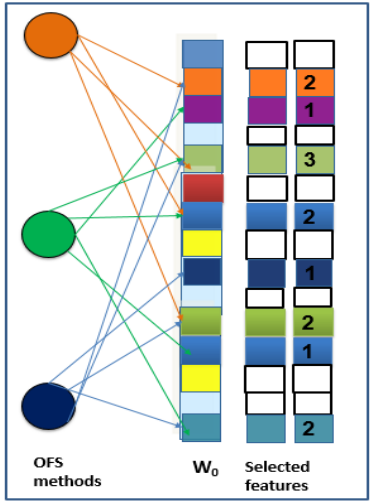

Fig. 6. Trust of selected features

where the numbers 1, 2 and 3 are the number of times a feature is selected. In this example (Fig. 6), just three OFS methods are present in the negotiation process, the vector $w_0$ contains $n$ features or weight of features and the vector *selected features* is a sparse vector with two layers where the first layer contains weights of different features (non-selected features has 0 as value and selected features has a weight different from 0). The second one is the layer of feature trust which contains the value of trust of each selected feature. These theoretical ideas should be validated experimentally. Next, we will present several experiments using the novel negotiation based OFS method.

### 3.3 ANOFS: Experimental Results

*Experimental Testbed and compared algorithm.* We conduct many experiments on a variety of benchmark datasets from UCI machine learning repository [40] and LIBSVM website. We have chosen these datasets arbitrarily in order to cover various sizes of them. Besides these datasets, we apply our proposed ANOFS methods (BANOFS and MANOFS) to real text classification datasets: (i) the Reuters Corpus Volume 1 (RCV1); (ii) 20 Newsgroups. In fact, we use a tiny version of the 20newsgroups data, with binary data for 100 words across 16242 postings, where we extract the "rec" versus "sci" and "comp" versus "rec" to form two binary classification tasks.

Table 1 shows details of our experimental testbed.

We compare our algorithm Multilateral Automated Negotiation based OFS to four algorithms RAND, PEtrun, OFS and BANOFS.

- "RAND": the randomized feature selection algorithm;
- "PEtrun": the modified perceptron by a simple truncation algorithm [10];
- "OFS": the online feature selection based sparse online learning algorithm [10].
- "BANOFS": our Bilateral ANOFS method proposed in [41].

In order to make an equitable comparison, all algorithms adopt the same experimental settings.

TABLE 1
LIST OF DATASETS USED IN ANOFS FRAMEWORK EXPERIMENTS

| Dataset | # Samples | # Dimensions |
|---|---:|---:|
| a9a | 32561 | 123 |
| covtype | 581012 | 54 |
| gisette | 6500 | 5000 |
| ijcnn1 | 49990 | 22 |
| Kddcup08 | 102264 | 117 |
| magic | 19020 | 10 |
| mushrooms | 8124 | 112 |
| spambase | 4601 | 56 |
| w8a | 64700 | 300 |
| rcv1 | 20242 | 47236 |
| 20News1 | 2000 | 1355191 |
| 20newsgroup("rec"vs"sci") | 6176 | 100 |
| 20newsgroup ("comp"vs"rec") | 8124 | 100 |

For every dataset, the number of selected features is B=round(0.1*dimensionality).
All the experiments were conducted over 20 times.

*Performance Evaluation.* We evaluate the online predictive performance and CPU running time of MANOFS and the fourth compared algorithms (as the average number of mistakes) on several data in Table 2 and on real text classification datasets in Table 3. We observe a considerable enhancement in the accuracy of MANOFS method compared to BANOFS and to the other related methods. In fact, this result is explained since MANOFS method uses a trust model which allows each negotiator to learn more from the history of feature selection system. This consequently forms experience about which feature is so important and which feature is so irrelevant and so on. Also, the number of negotiators can make the difference and can explain these results. In fact, these learners-negotiators are cooperative so the skill of each participant helps the improvement of feature selection process.

More specifically, using the MANOFS method makes us think about a multi-agent system with a negotiation process that resolves the problem of online feature selection.

This idea was implemented and tested and we will describe our OFS system in the next section.

## 4 MULTI-OBJECTIVE AUTOMATED NEGOTIATION BASED ONLINE FEATURE SELECTION SYSTEM

In this section, we will present our OFS system as Multi Agent System (MAS) named **MOANOFS** which resolves multi-issue negotiation based OFS.

Let's note that in the work described before [39], [41], [42] we resolved a simple-issue negotiation based OFS; this means our objective was to minimize the error rate of prediction even with two learners or more.

Next, we will present our challenge and our idea to get our OFS system. Then, we will describe our MOANOFS system. Finally, we will show some experiment results using multi-issue negotiation based Online Feature Selection system (MOANOFS) compared to our simple-issue negotiation based OFS system (MALOFS) [39], [41] and other state of the art OFS methods.

TABLE 2
EVALUATION OF THE AVERAGE NUMBER OF MISTAKES AND CPU RUNNING TIME (S) BY FIVE ALGORITHMS ON EIGHT DATASETS

| Algorithm | a9a | Covtype |
|---|---|---|
| RAND | 19051.1 ± 0102.3 (0.953) | 351083.1 ± 0287.7 (14.481) |
| PEtrun | 11949.5 ± 0070.2 (0.949) | 277676.2 ± 2023.8 (17.259) |
| OFS | 7373.4 ± 0389.6 (1.084) | 290351.3 ± 0307.1 (18.803) |
| BANOFS | 7079.4 ± 0054.0 (**0.584**) | 272908.3 ± 0317.1 (12.736) |
| **MANOFS** | **1913.1 ± 0427.8 (0.863)** | **64846.1 ± 1450.0 (08.062)** |
| Algorithm | ijcnn1 | Kddcup08 |
| RAND | 20338.1 ± 0196.3 (0.839) | 50898.1 ± 00159.7 (2.893) |
| PEtrun | 40056.1 ± 0086.6 (1.955) | 59562.6 ± 00358.6 (4.456) |
| OFS | 39177.8 ± 8075.4 (2.114) | 58199.5 ± 00316.4 (5.379) |
| BANOFS | 5287.1 ± 0039.1 (**0.521**) | 43280.8 ± 18655.1 (2.869) |
| **MANOFS** | **4152.1 ± 0928.4 (0.812)** | **11305.7 ± 02528.0 (2.416)** |
| Algorithm | mushrooms | Spambase |
| RAND | 4054.1 ± 034.7 (0.108) | 1827.9 ± 039.2 (0.084) |
| PEtrun | 4052.3 ± 033.8 (0.155) | 1294.8 ± 066.3 (0.090) |
| OFS | 4052.3 ± 033.8 (0.164) | 913.1 ± 157.8 (0.170) |
| BANOFS | 3223.1 ± 121.0 (0.097) | 589.7 ± 013.7 (**0.054**) |
| **MANOFS** | **915.6 ± 204.7 (0.087)** | **167.7 ± 037.5 (0.071)** |
| Algorithm | gisette | w8a |
| RAND | 3013.0 ± 39.7 (2.579) | 56257.1 ± 0074.6 (2.194) |
| PEtrun | 544.5 ± 26.9 (1.266) | 55249.6 ± 2086.9 (2.515) |
| OFS | 3014.6 ± 39.4 (3.526) | 18499.5 ± 5203.0 (1.351) |
| BANOFS | 452.4 ± 11.0 (**1.071**) | 7648.0 ± 0032.3 (**0.965**) |
| **MANOFS** | **124.7 ± 27.9 (2.280)** | **4471.6 ± 0999.9 (1.253)** |

TABLE 3
EVALUATION OF THE AVERAGE NUMBER OF MISTAKES BY FIVE ALGORITHMS ON REAL TEXT CLASSIFICATION DATASETS

| Algorithm | RCV1 | 20news1 |
|---|---|---|
| RAND | 16224.2 ± 50.7 (159.3) | 1165.6 ± 24.1 (126.0) |
| PEtrun | 1448.0 ± 27.4 (**120.0**) | 141.8 ± 05.9 (0**22.8**) |
| OFS | 1496.6 ± 23.2 (194.3) | 191.3 ± 31.3 (164.6) |
| BANOFS | 1375.1 ± 20.0 (124.9) | 141.8 ± 05.9 (034.3) |
| **MANOFS** | **326.9 ± 73.1 (411.0)** | **36.0 ± 08.0 (161.7)** |
| Algorithm | 20newsgroup ("rec"vs"sci") | 20newsgroup ("comp"vs"rec") |
| RAND | 6070.5 ± 011.2 (0.231) | 7922.7 ± 017.0 (0.338) |
| PEtrun | 4473.4 ± 070.7 (0.267) | 5356.9 ± 102.2 (0.378) |
| OFS | 4112.0 ± 198.5 (0.284) | 4370.2 ± 238.1 (0.357) |
| BANOFS | 1351.6 ± 017.7 (**0.100**) | 1165.5 ± 015.8 (**0.128**) |
| **MANOFS** | **1006.2 ± 225.0 (0.121)** | **1240.1 ± 277.3 (0.162)** |

## 4.1 Challenge

First of all, we mention wherefrom the idea of designing such system is derived. In fact, we remark that distinguishing relevant features from redundant ones, in an ultra-high dimensional and massive training dataset and in an online fashion, is a big challenge. It can even be considered as a decision-making problem. In this paper, we overcame this challenge by developing an intelligent online feature selection system based on the principle of automated negotiation. We have designed an Online Feature Selection System using multiple collaborative agent learners. Several intelligent agent learners were used to improve the online classifier performance.

In fact, the idea behind this system was inspired by multi-classifiers systems proposed by Hamdani *et al.* [26] in which the classification performance is enhanced using interaction and communication concepts. In our case, we are interested in the analysis step in a pattern recognition system. Our aim is to enhance the decision-making system by improving the feature selection technique. Likewise, our OFS system is a new vision of OFS literature systems. We consider the online feature selection methods as intelligent agents. We suppose that we are in the context of decision making in order to resolve the problem of; which are relevant features? In the same context, we also integrate the principle of negotiation between agent-learners to have a multi-agent system (MAS) that supports machine learning model to reflect the whole complexity of the real world. Below, we present in details our OFS system.

## 4.2 MOANOFS: Description

We suppose that our negotiation space is represented by $Neg = \langle P, A, D, U, T \rangle$.

**P** represents the set of participants where we have three OFS methods as participants in the negotiation process.

**A** = {$a_1$, $a_2$, $a_3$} represents the set of attributes or issues where $a_1$ = *Trust*, $a_2$ = *ErrorRate* and $a_3$ = *TimeCost*. Each attribute $a_i$ takes its value from the corresponding domain $D_{ai}$.

In a negotiation round, each OFS-Neg will make an offer $o$ and send it to the initiator: $\vec{o} = \langle d_{a_1}, d_{a_2}, d_{a_3}\rangle$ where $d_{a_1} = Max_{Trust}, d_{a_2} = Min_{Error}, and\ d_{a_3} = Min\_CostTime$.

The valuation function for each OFS-Neg $p$ for each attribute $a \in A$ is defined by: $U_p^A: A \to [0,1]$. The valuations of attributes are assumed normalized:

$$U_p(Trust) + U_p(Error) + U_p(CostTime) = 1.$$

In fact, this is an example to explain more our process: each participant $p$ in the negotiation process MANOFS aims first to minimize the error rate of prediction and then of classification. Second, it tends to minimize the time cost of the feature selection cost-time. Finally, our MOANOFS system has as third object to maximize its trust in this system. However, in any negotiation process, each participant gives value to each objective; this means a negotiator p can give more importance to the first object than the second and so one. A negotiator p is interested to maximize Trust objective by 0.2 ($U_p(Trust) = 0.2$), it gives importance to minimizing the error rate and the cost-time objectives by 0.5 ($U_p(Error) = 0.5$) and 0.3 ($U_p(CostTime) = 0.3$) respectively. This example shows that the sum of these three objectives values is 1.

The valuation function of each OFS-Neg $p$ for each attribute value $d_a \in D_a$ is defined by: $U_p^{D_a}: D_a \to [0,1]$.

To measure the utility function of an offer o, we use a common way as following:
$$U_p^o(o) = U_p(Trust) \times U_p(Max\_Trust) + U_p(Error) \\ \times U_p(Min\_Error) + U_p(CostTime) \\ \times U_p(Min\_CostTime)$$

where $U_p(Min\_Error)$ for example is the valuation of the objective 'Minimization of Error rate of prediction' which means the offer in Error Rate given by an OFS-Neg $p$.

It is clear that we are instead of a multi-objective problem. Our objective is minimizing this utility function in the case of transforming the maximization of trust by minimizing (1- trust) since trust should be in [0,1]. But, if we have two different types of objectives, in other words, if we should minimize one object and maximize another, we will have a complex multi objective problem. To resolve such problem, we propose to integrate a multi-objective particle swarm optimization MOPSO in each agent-learner. For each learner-negotiator (OFS-Neg), a population of particles is used to represent the offers. Each particle is composed of three issues as shown in Fig. 7.

| ID | Error | CostTime | Trust |
|----|-------|----------|-------|

Fig. 7. OFS-Neg particle

This idea can be applied in our system but in the experimental results, we treat the case of minimizing the utility function with three issues. Next, we will present the section of experimental results using MOANOFS system.

The principle component of our MOANOFS system is the intelligent Agent-Learner (the OFS method).

In our OFS system, we developed a framework of online feature selection based on first-order and second-order online learning methods [44]. We use LIBOL [38] which is a library of online learning methods as machine learning tool. The different OFS algorithms based LIBOL are listed in [38]. In fact, these OFS algorithms are our Agents-Learners.

**OFS-Neg** (**OFS Neg**otiator) is defined from two big domains. The first is the Feature Selection domain where OFS-Neg is an OFS method. OFS-Neg is a generalization of a First-order and Second-order Online Learning based truncation as Feature Selection technique. The second domain is the multi-agent systems where OFS-Neg is an agent that benefits from the characteristics of an agent in a MAS such as negotiation and environment adaptation and evolution (with *Trust model*).

Trust mechanism is usually used to help negotiators to select whom they should interact with. Trust is formed from one's own past experiences with other negotiators. To put it simpler, trust is the power of each negotiator building on its history.

Reputation is built according to the opinion that agent societies have on individuals. Trust and reputation mechanisms may help to reduce conflict by interacting with good partners.

Anupam and Mohamed presented in their paper [43] different types of trust model applied in multi-agent systems. Example, *Direct Trust model* or which is known as local trust. This type of trust represents the portion of trust that an agent computes from its own experience about another agent (called target agent).

Let $DT_n^t(p1, p2)$ represents the direct trust that agent $p_1$ has upon agent $p_2$ up to $n$ transactions in the $t$th time interval. We have used the satisfaction measure to define direct trust as follows:
$$DT_n^t(p_1, p_2) = Sat_n^t(p_1, p_2) \quad (12)$$

where $Sat_n^t(p_1, p_2)$ represents the amount of satisfaction agent $p_1$ has upon agent $p_2$ based on its service up to $n$ transactions in the $t$th time interval. The initial value of satisfaction is $Sat_0^0(p_1, p_2) = 0$.

$$Sat_n^t(p_1, p_2) = \alpha \times Sat_{cur} + (1 - \alpha) \times Sat_{n-1}^t(p_1, p_2) \quad (13)$$

Where:
- ✓ $Sat_{cur}$ represents the satisfaction value for the most recent transaction where $Sat_{cur} = 0$ if transaction is fully unsatisfactory; $Sat_{cur} = 1$ if transaction is fully satisfactory; and $Sat_{cur} \in (0,1), otherwise$.
- ✓ $\alpha$ changes based on the accumulated deviation $\xi_n^t(p_1, p_2)$.
$$\alpha = threshold + c \times \frac{\delta_n^t(p_1, p_2)}{1 + \xi_n^t(p_1, p_2)}, \quad (14)$$
$$\delta_n^t(p_1, p_2) = |Sat_{n-1}^t(p_1, p_2) - Sat_{cur}|, \quad (15)$$
$$\xi_n^t(p_1, p_2) = c \times \delta_n^t(p_1, p_2) + (1 - c) \times \xi_{n-1}^t(p_1, p_2). \quad (16)$$
- ✓ $\xi_0^t(p_1, p_2) = \xi_{last}^{t-1}(p_1, p_2)$ and $\xi_0^0(p_1, p_2) = 0$.
- ✓ $c$ is some user defined constant factor which controls to what extent we will react to the recent error $\delta_n^t(p_1, p_2)$. If we increase the value of $c$, we will give more significance to the recent deviation than the accumulated deviation and vice versa.
- ✓ The *threshold* represents a threshold which is used to prevent $\alpha$ from saturating to a fixed value. Initial value of $\alpha$ is set to 1 and *threshold* is set to 0.25.

The principal objective of [43] is to provide a dynamic trust computation model which evaluates the trust of agents even in the presence of highly oscillating malicious behavior.

In fact, Trust can be used in various applications such as Spam Filtering, Recommender Systems, P2P File Sharing, etc.

Our MOANOFS system (see Fig. 8) contains two levels of selection. In the first level, the more confident OFS-Neg is selected (so the selection is between agents). In the second level, relevant features are selected using our negotiation method MANOFS (so the selection is between features).

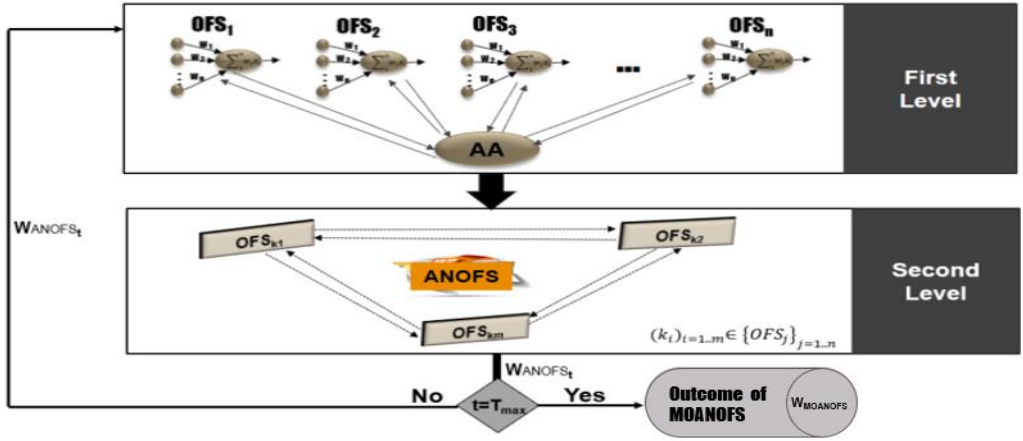

Fig. 8. MOANOFS: the proposed multi-objective OFS System

## 4.3 MOANOFS: Experimental Results

In this section, we show results of extensive experiments conducted to evaluate our OFS system (MOANOFS).

### A. Experimental Testbed

We evaluate our OFS system on a number of benchmark datasets from web machine learning repositories in order to examine the binary classification performance. Table 4 shows the details of all datasets in our experiments (you can find details on these datasets in [39]).

Some datasets can be downloaded from UCI machine leaning repository and LIBSVM website. We also apply our proposed MOANOFS to real text classification datasets: 20 Newsgroups[1] datasets. In fact, we have chosen to cover various sizes of datasets to conduct experiments on medium-scale and large-scale data, also on high dimensional large-scale real text and web data classification.

### B. Compared Algorithms

MOANOFS to our simple-issue negotiation based OFS system (MALOFS) [39], [41] and to other state of the art OFS methods [10], [18], [19][44]. Table 5 lists details of the compared OFS algorithms. All these algorithms have been developed for binary classification tasks. The experiments are conducted over 10 times with a random permutation of datasets. We exploit the OL's advantages which processes sequentially the data.

For every dataset, and as a first experiment, we set the number of selected features as *round(0.1\*dimensionality)*. Then, in a second experiment, we vary the number of selected features from one OFS-Neg to another (10%, 20%, and 30%).

In the experiments below, we show the importance of our OFS system compared to the execution of each OFS algorithm alone even by varying the B parameter (see Table 6) or the loss function between the different OFS-Neg.

TABLE 4
DETAILS OF DATASETS USED IN MOANOFS EXPERIMENTS

| Several Large-Scale Benchmark Dataset | | |
|---|---|---|
| **Dataset** | **# Samples** | **# Dimensions** |
| a9a | 32561 | 123 |
| covtype | 581012 | 54 |
| gisette | 6000 | 5000 |
| ijcnn1 | 49990 | 22 |
| Kddcup08 | 102294 | 117 |
| mushrooms | 8124 | 112 |
| spambase | 4601 | 56 |
| w8a | 64700 | 300 |
| a8a | 32561 | 123 |
| **Large-Scale Real Dataset** | | |
| **Dataset** | **# Samples** | **# Dimensions** |
| Pcmac | 1943 | 7510 |
| relathe | 1427 | 4322 |
| basehock | 1993 | 4863 |
| 20newsgroup("rec"vs"sci") | 6176 | 100 |
| 20newsgroup("comp"vs "rec") | 8124 | 100 |

In our experiments, we compare the proposed

In fact, each OFS-Neg chooses a loss type to work with; example *AROW-OFS* uses hinge loss as the loss function, but *PA-OFS, SOP-OFS* and *ROMMA-OFS* choose zero-one loss function, etc.

First of all, we present results of MOANOFS on several datasets by varying the number of selected features $B$.

### C. Experiment I: Evaluation of MOANOFS by Varying B on Several Data

In this experiment, we choose randomly some datasets to test the influence of the $B$ variation; the number of selected features; on the efficiency of our online feature selection system.

We conducted this experiment by fixing three times the value of B (10%, 20% and 30%) and at a fourth time, we vary it randomly around the different OFS-Neg (eg. OFS-Neg1 selects 10%, OFS-Neg2 selects 20% and OFS-Neg3 selects 30%).

[1] http://qwone.com/~jason/20Newsgroups/

TABLE 5
LIST OF COMPARED ALGORITHMS

| OFS Algorithm | Description | Category |
|---|---|---|
| PET | Perceptron based OFS [10] | 1st Order |
| ROMMA-OFS | Relaxed Online Maximum Margin Algorithm based OFS [44] | 1st Order |
| ALMA-OFS | Approximate Large Margin algorithm based OFS [44] | 1st Order |
| OGD-OFS | Online Gradient Descent based OFS [44] | 1st Order |
| PA-OFS | Passive Aggressive based OFS [44] | 1st Order |
| SOP-OFS | Second Order Perceptron based OFS [44] | 2nd Order |
| CW-OFS | Confidence Weighted based OFS [18][19] | 2nd Order |
| AROW-OFS | Adaptive Regularization of Weight vectors based OFS [18], [19] | 2nd Order |
| SCW-OFS | Soft Confidence-Weighted learning algorithm based OFS [44] | 2nd Order |
| MALOFS | Multi Agent-Learners based OFS [39], [41] | 1st & 2nd Order |
| **MOANOFS** | Multi-Objective Automated Negotiation based OFS (our proposed OFS system in this paper) | 1st & 2nd Order |

TABLE 6
EVALUATION OF MOANOFS (PREDICTION ERROR RATE) ON SEVERAL DATASETS BY VARYING B

| | 20newsgroup ("rec" vs "sci") | 20newsgroup ("comp" vs "rec") |
|---|---|---|
| B = 10% | 2.91 ± 0.05 (4.27s) | 1.33 ± 0.03 (1.09s) |
| B = 20% | 4.32 ± 0.06 (2.21s) | 2.30 ± 0.03 (1.20s) |
| B = 30% | 6.01 ± 0.06 (3.01s) | 3.32 ± 0.03 (1.33s) |
| B varied | 3.26 ± 0.05 (2.20s) | 1.70 ± 0.03 (1.09s) |
| | **a9a** | **spambase** |
| B = 10% | 3.49 ± 0.06 (06.84s) | 3.75 ± 0.05 (0.56s) |
| B = 20% | 6.99 ± 0.07 (10.19s) | 5.81 ± 0.05 (0.44s) |
| B = 30% | 9.65 ± 0.07 (11.39s) | 6.97 ± 0.05 (0.62s) |
| B varied | 3.89 ± 0.06 (05.60s) | 5.57 ± 0.05 (0.48s) |
| | **ijcnn1** | **mushrooms** |
| B = 10% | 1.58 ± 0.03 (9.72s) | 07.24 ± 0.10 (0.91s) |
| B = 20% | 2.22 ± 0.03 (9.26s) | 16.49 ± 0.17 (0.91s) |
| B = 30% | 2.84 ± 0.03 (3.78s) | 48.20 ± 2.27 $e^{-16}$ (0.58s) |
| B varied | 4.39 ± 0.03 (3.55s) | 48.20 ± 2.27 $e^{-16}$ (0.60s) |

Table 6 shows different results (prediction error rate (%) and time cost (second s)) of MOANOFS by varying the selected number of features.

We observe that when we fix the number of selected features, the error rate as well as the time-cost increase as we select more features. This can be explained by the efficiency of every online feature selection model by selecting minimum features and of course they should be the relevant ones to keep the meaning of the data in case.

The idea of varying $B$ in different agent-learners of our MOANOFS was inspired by the importance of being different in a decision-making system. This difference can lead to better results. However, in our case, we observe that this variation enhances the CPU time_cost but it also has no interesting accuracy results.

If our objective is simply *minimizing the CPU time_cost*, the *varied B* will be a good choice.

We observe that 10% is the best choice if our objective is to minimize the error rate of predictors. For this reason, we choose to work with B = 10% in the following experiments.

### D. Experiment II: Evaluation of MOANOFS on Several Large-Scale Data

In this experiment, we evaluate the performance of online feature selection algorithms on a number of public large-scale benchmark datasets as shown in Table 6.

Table 7 presents the online predictive performance of the compared algorithms by fixing the number of selected features to 10% of the dataset dimension. Several observations can be conducted. First of all, we discuss the result of the first level which is the selection of the trustful OFS-Neg. In fact, the three elected learners differ from a test to another.

This means, that the system remarkably tries to choose the OFS-Neg which not only has the better accuracy but also which is confident all over the experiment. We make the result of the elected OFS-Neg in blue color. The MOANOFS result is in bold and in blue color too.

We observe that MOANOFS's performance is stable. The variation of MOANOFS is not larger than those of the other online feature selection algorithms on all over the datasets. The proposed MOANOFS is able to learn a more compact classification model. With the same number of selected features, MOANOFS is also able to achieve higher accuracy. However, we observe that our system can't enhance the time-cost in all datasets. But regarding to the balance between the minimum error of prediction and the CPU time-cost, and considering that we have the 'covtype' dataset as example, the variance is not high between neither the time of MOANOFS nor that of SCW since MALOFS outperforms in accuracy (MOANOFS: 1-6.9 % and SCW-OFS: 1-23.3%).

### E. Experiment III: Evaluation of MOANOFS on Large-Scale Real Data

In this experiment, we evaluate our system on a number of large scale public benchmark real datasets which can be downloaded from SVMLin[2]. The results are shown in

Table 8.

These experiments on real datasets support our observations done before: our OFS system enhances the accuracy of prediction and this is due to the cooperation between multiple learners that have different preferences.

---
[2] http://vikas.sindhwani.org/svmlin.html

TABLE 7
EVALUATION OF MOANOFS ON NINE LARGE-SCALE BENCHMARK DATASETS (B=10%)

| Algorithm | a9a | Covtype |
|---|---|---|
| PET | 21.0 ± 0.10 (1.61s) | 47.0 ± 0.10 (026.5s) |
| ROMMA-OFS | 22.5 ± 0.30 (1.64s) | 47.8 ± 0.90 (028.1s) |
| ALMA-OFS | 15.8 ± 0.10 (1.83s) | 48.4 ± 0.10 (030.7s) |
| OGD-OFS | 15.6 ± 0.10 (**1.76s**) | 46.7 ± 0.10 (0**31.8s**) |
| PA-OFS | 21.1 ± 0.10 (1.77s) | 48.4 ± 0.00 (030.2s) |
| SOP-OFS | 20.9 ± 0.10 (6.17s) | 33.7 ± 0.00 (048.7s) |
| CW-OFS | 21.1 ± 0.20 (4.76s) | 48.7 ± 3.80 (229.2s) |
| AROW-OFS | 15.4 ± 0.10 (4.04s) | 24.4 ± 0.00 (052.5s) |
| SCW-OFS | 15.5 ± 0.10 (3.06s) | 23.3 ± 0.10 (047.7s) |
| MALOFS | 15.4 ± 0.00 (4.10s) | 23.4 ± 0.00 (046.0s) |
| **MOANOFS** | **03.4 ± 0.06 (3.21s)** | **06.9 ± 0.10 (045.7s)** |
| Algorithm | Spambase | ijcnn1 |
| PET | 47.4 ± 0.7 (0.30s) | 10.6 ± 0.10 (1.95s) |
| ROMMA-OFS | 29.0 ± 2.1 (0.31s) | 10.1 ± 0.10 (1.96s) |
| ALMA-OFS | 47.9 ± 0.5 (0.35s) | 07.4 ± 0.10 (**2.25s**) |
| OGD-OFS | 44.6 ± 0.6 (0.36s) | 09.5 ± 0.10 (2.42s) |
| PA-OFS | 33.1 ± 0.4 (0.32s) | 10.3 ± 0.10 (2.07s) |
| SOP-OFS | 16.4 ± 0.6 (0.55s) | 10.4 ± 0.10 (3.42s) |
| CW-OFS | 13.4 ± 0.4 (0.50s) | 10.1 ± 0.10 (3.06s) |
| AROW-OFS | 09.4 ± 0.4 (0.53s) | 07.8 ± 0.10 (3.18s) |
| SCW-OFS | 10.6 ± 0.2 (0.43s) | 06.7 ± 0.10 (2.65s) |
| MALOFS | 10.5 ± 0.1 (0.61s) | 06.6 ± 0.11 (5.25s) |
| **MOANOFS** | **03.7 ± 0.2 (0.27s)** | **01.5 ± 0.03 (2.49s)** |
| Algorithm | w8a | Gisette |
| PET | 11.4 ± 0.06 (06.05s) | 07.6 ± 0.20 (0005.4s) |
| ROMMA-OFS | 10.8 ± 0.02 (06.27s) | 10.4 ± 0.70 (0005.3s) |
| ALMA-OFS | 10.0 ± 0.06 (06.65s) | 50.2 ± 0.70 (0005.4s) |
| OGD-OFS | 10.2 ± 0.04 (07.00s) | 07.4 ± 0.30 (0005.1s) |
| PA-OFS | 10.4 ± 0.02 (06.26s) | 05.1 ± 0.20 (0005.0s) |
| SOP-OFS | 11.6 ± 0.06 (40.00s) | 46.6 ± 0.70 (3002.7s) |
| CW-OFS | 10.2 ± 0.03 (14.00s) | 03.8 ± 0.10 (0577.3s) |
| AROW-OFS | 09.8 ± 0.03 (23.10s) | 05.1 ± 0.30 (4974.5s) |
| SCW-OFS | 09.5 ± 0.02 (12.70s) | 03.8 ± 0.10 (0663.9s) |
| MALOFS | 09.4 ± 0.01 (14.01s) | 03.6 ± 0.00 (0960.1s) |
| **MOANOFS** | **02.1 ± 0.04 (12.80s)** | **00.3 ± 0.01 (0744.5s)** |
| Algorithm | Kddcup08 | Mushrooms |
| PET | 49.80 ± 00.2 (05.4s) | 39.80 ± 1.2 (0.34s) |
| ROMMA-OFS | 45.80 ± 01.1 (05.7s) | 39.20 ± 8.4 (0.35s) |
| ALMA-OFS | 49.40 ± 00.2 (06.2s) | 49.90 ± 0.4 (0.39s) |
| OGD-OFS | 42.50 ± 00.8 (06.6s) | 34.20 ± 1.2 (**0.34s**) |
| PA-OFS | 38.50 ± 00.1 (06.1s) | 47.60 ± 0.4 (0.38s) |
| SOP-OFS | 48.40 ± 01.3 (20.4s) | 09.50 ± 0.3 (0.55s) |
| CW-OFS | 44.40 ± 10.0 (16.4s) | 10.68 ± 0.3 (0.50s) |
| AROW-OFS | 41.20 ± 03.0 (21.2s) | 16.60 ± 0.2 (0.53s) |
| SCW-OFS | 15.40 ± 08.3 (17.1s) | 43.60 ± 1.1 (0.43s) |
| MALOFS | 15.40 ± 00.0 (20.5s) | 05.90 ± 1.1 (0.40s) |
| **MOANOFS** | **07.24 ± 00.1 (00.6s)** | **07.20 ± 0.1 (0.61s)** |

TABLE 8
EVALUATION OF MOANOFS ON REAL DATASETS (B=10%)

| Algorithm | Pcmac | Basehock |
|---|---|---|
| PET | 19.3 ± 1.40 (001.0s) | 07.6 ± 0.40 (001.3s) |
| ROMMA-OFS | 19.8 ± 1.30 (001.1s) | 12.2 ± 1.00 (001.5s) |
| ALMA-OFS | 17.1 ± 1.30 (001.1s) | 36.6 ± 1.30 (001.6s) |
| OGD-OFS | 17.9 ± 1.90 (001.1s) | 07.7 ± 1.50 (001.5s) |
| PA-OFS | 14.4 ± 0.50 (001.1s) | 05.2 ± 0.40 (001.6s) |
| SOP-OFS | 36.4 ± 0.90 (139.5s) | 33.6 ± 0.90 (383.9s) |
| CW-OFS | 11.0 ± 0.30 (085.8s) | 02.9 ± 0.20 (210.6s) |
| AROW-OFS | 12.6 ± 0.40 (092.8s) | 06.4 ± 0.50 (178.1s) |
| SCW-OFS | 11.0 ± 0.30 (083.5s) | 03.1 ± 0.20 (165.2s) |
| MALOFS | 10.0 ± 0.10 (090.1s) | 03.0 ± 0.10 (170.3s) |
| **MOANOFS** | **0.05 ± 0.00 (071.4s)** | **00.1 ± 0.00 (069.2s)** |
| Algorithm | 20newsgroup ("rec" vs "sci") | 20newsgroup ("comp" vs "rec") |
| PET | 17.5 ± 0.30 (0.29s) | 11.6 ± 0.2 (0.38s) |
| ROMMA-OFS | 18.3 ± 0.30 (0.29s) | 11.8 ± 0.2 (0.38s) |
| ALMA-OFS | 14.1 ± 0.20 (0.33s) | 08.9 ± 0.2 (0.44s) |
| OGD-OFS | 14.4 ± 0.20 (0.35s) | 08.9 ± 0.20 (0.46s) |
| PA-OFS | 18.4 ± 0.30 (0.31s) | 11.9 ± 0.20 (0.40s) |
| SOP-OFS | 19.3 ± 0.30 (0.93s) | 13.2 ± 0.30 (1.25s) |
| CW-OFS | 17.6 ± 0.30 (0.76s) | 11.2 ± 0.20 (0.91s) |
| AROW-OFS | 13.7 ± 0.30 (0.84s) | 08.5 ± 0.20 (1.07s) |
| SCW-OFS | 13.6 ± 0.20 (0.74s) | 08.4 ± 0.10 (0.89s) |
| MALOFS | 13.5 ± 0.00 (2.01s) | 08.3 ± 0.00 (3.50s) |
| **MOANOFS** | **02.9 ± 0.05 (0.57s)** | **01.3 ± 0.03 (0.68s)** |

## 5 CONCLUSION

We have developed an intelligent online feature selection system based on automated negotiation principle. In this system, we designed an Online Feature Selection System using multiple collaborative agent learners. Actually, we used multiple agent learners to implement our OFS system.

Indeed, the idea of this work was inspired by the multi-classifier systems as in [26] where they use interaction and communication concepts to enhance the classification performance.

In our case, we are interested in the step of analysis in a pattern recognition system. Our aim is to improve the feature selection system in order to enhance the decision-making system. Likewise, our OFS system is a new vision of OFS literature systems. We considered the online learning methods as intelligent agents. We supposed that we are in the context of decision making problem to resolve the problem of how to distinguish the relevant features. In addition, we integrated, in the same context, the principle of negotiation between the agent-learners to have a multi-agent system that supports machine learning model to reflect the whole complexity of the real world. We have used a multi-objective negotiation technique where we tend to resolve the problem of online feature selection by minimizing the error and the time-cost of prediction, and maximizing the trust of each agent-learner. This leads to a multi-issue negotiation based online feature selection system MOANOFS.

MOANOFS is successfully applicable to different do-mains and it achieves high accuracy with some real-world applications.

Our work may be extended to resolve the problem of online feature selection for multi-classification domains. Also, we will resolve the problem of online feature selection in nonlinear domains, with unsupervised learning methods and unsupervised classification from stream data: a more real and complex problem.

## ACKNOWLEDGMENT

The research leading to these results has received funding from the Ministry of Higher Education and Scientific Research of Tunisia under the grant agreement number LR11ES48.

## REFERENCES

[1] L. C. Molina, L. Belanche, and A. Nebot, "Feature selection algorithms: A survey and experimental evaluation," *Proc. IEEE International Conference on Data Mining (ICDM '02)*, pp. 306-313, Dec. 2002, doi:10.1109/ICDM.2002.1183917.

[2] J. Tang, and H. Liu, "Feature selection with linked data in social media," *Proc. Twelfth SIAM International Conference on Data Mining (SDM'12)*, pp. 118-128, Apr. 2012, doi:10.1137/1.9781611972825.11.

[3] H. Liu and L. Yu, "Toward integrating feature selection algorithms for classification and clustering," *IEEE Trans. on Knowledge and Data Engineering*, vol. 17, no. 4, pp. 491-502, March 2005, doi:10.1109/TKDE.2005.66.

[4] M. Dash and H. Liu, "Feature selection for classification," *Intelligent data analysis*, vol. 1, no. 3, pp. 131-156, 1997.

[5] B. Ghaddar, and J. Naoum-Sawaya, "High Dimensional Data Classification and Feature Selection using Support Vector Machines," *European Journal of Operational Research*, vol. 265, no. 3, pp. 993-1004, March 2018.

[6] Z. Zhao, L. Wang, H. Liu, and J. Ye, "On similarity preserving feature selection," *IEEE Trans. on Knowledge and Data Engineering*, vol. 25, pp. 619–632, March 2013, doi:10.1109/TKDE.2011.222.

[7] S. Perkins, and J. Theiler, "Online feature selection using grafting," *Proc. 20th International Conference on Machine Learning (ICML'03)*, pp. 592–599, 2003.

[8] X. Wu, K. Yu, H. Wang, and W. Ding, "Online streaming feature selection", *Proc. of the 27th International conference on machine learning (ICML'10)*, pp. 1159-1166, 2010.

[9] X. Wu, K. Yu, W. Ding, H. Wang, and X. Zhu, "Online Feature Selection with Streaming Feature," *IEEE Trans. Pattern Analysis and Machine Intelligence*, vol. 35, no. 5, May 2013, doi:10.1109/TPAMI.2012.197.

[10] J. Wang, P. Zhao, S. C.H. Hoi, and R. Jin, "Online Feature Selection and Its Applications", *IEEE Trans. Knowledge and Data Engineering (TKDE)*, vol. 26, pp. 698-710, March 2014, doi:10.1109/TKDE.2013.32.

[11] J. Li, Y. Cao, Y. Wang, and H. Xiao, "Online Learning Algorithms for Double-Weighted Least Squares Twin Bounded Support Vector Machines," *Neural Processing Letters*, vol. 43, no. 1, pp. 1-21, May 2016.

[12] K.C. Lee and D. Kriegman, "Online learning of probabilistic appearance manifolds for video-based recognition and tracking," in IEEE Computer Society Conference on Computer Vision and Patttern Recognition CVPR, San Diego, CA, USA, 20-26 June 2005, pp. 852-859.

[13] K. Hara, M. Okada, "Online learning theory of ensemble learning using linear perceptrons," in IEEE International Joint Conference on Neural Networks IJCNN, Budapest, Hungary, 25-29 July 2004, vol. 2, pp. 1139-1143,

[14] K. Crammer, O. Dekel, J. Keshet, S. Shalev-Shwartz, and Y. Singer, "Online Passive-Aggressive Algorithms", *Journal of Machine Learning Research (JMLR)*, 2006.

[15] P. Zhao, S. C. H. Hoi, and R. Jin, "Double Updating Online Learning", *Journal of Machine Learning Research*, vol. 12, pp. 1587-1615, 2011.

[16] N. R. Jennings, P. Faratin, A. R. Lomuscio, S. Parsons, M. J. Wooldridge, and C. Sierra, "Automated Negotiation: Prospects, Methods and Challenges," Group Decision and Negotiation (GDN), vol. 10, no. 2, pp. 199-215, 2001.

[17] J. Langford, L. Li, and T. Zhang, "Sparse online learning via truncated gradient," *Journal of Machine Learning Research (JMLR)*, vol. 10, pp. 777-801, Mar. 2009.

[18] Y. Wu, S. C. H. Hoi, and T. Mei, "Massive-scale Online Feature Selection for Sparse Ultra-high Dimensional Data," CoRR abs/1409.7794, Sept. 2014.

[19] Yue Wu, Steven C. H. Hoi, Tao Mei, Nenghai Y, "Large-Scale Online Feature Selection for Ultra-High Dimensional Sparse Data," ACM Transactions on Knowledge Discovery from Data (ACM TKDD), vol. 11, no. 4, pp. 48:1-48:22, June 2017.

[20] K. Yu, X. Wu, W. Ding, and J. Pei1, "Towards Scalable and Accurate Online Feature Selection for Big Data," in IEEE International Conference on Data Mining (ICDM 2014), Shenzhen, China, 14-17 December 2014, pp. 660-669.

[21] H. Yang, R. Fujimaki, Y. Kusumura, and J. Liu, "Online Feature Selection: A Limited-Memory Substitution Algorithm and Its Asynchronous Parallel Variation," in Proceedings of the 22nd ACM SIGKDD International Conference on Knowledge Discovery and Data Mining (KDD 2016), San Francisco, CA, USA, August 13-17, 2016, pp. 1945-1954.

[22] S. Perkins, K. Lacker, and J. Theiler, "Grafting: Fast, incremental feature selection by gradient descent in function space," Journal of Machine Learning Research (JMLR), vol. 3, pp. 1333-1356, Jan. 2003.

[23] J. Zhou, D. P. Foster, R. A. Stine, and L. H. Ungar, "Streamwise Feature Selection," Journal of Machine Learning Research (JMLR), vol. 7, pp. 1861-1885, Sept. 2006.

[24] G. Keerthika, and D. S. Priya, "Feature subset evaluation and classification using naive bayes classifier," Journal of Network Communications and Emerging Technologies (JNCET), vol.1, no. 1, pp. 22-27, Mar. 2015.

[25] J. Wang, M. Wang, P. Li, L. Liu, Z. Zhao, X. Hu, and X. Wu, "Online Feature Selection with Group Structure Analysis," IEEE Trans. on Knowledge and Data Engineering (TKDE), vol. 27, no. 11, pp. 3029-3041, June 2015.

[26] T. M. Hamdani, M. A. Khabou, A. M. Alimi, and F. Karray, "An Intelligent Decision-Making System Based on Multiple Classifiers Updated using Confidence Rates and Stress Parameters," Control and Intelligent Systems, vol. 39, no. 4, pp. 213-223, Dec. 2011.

[27] A. R. Lomuscio and N. R. Jennings, "A Classification Scheme for Negotiation in Electronic Commerce," Journal of Group Decision and Negotiation (GDN), vol. 12, no. 1, pp. 31-56, Jan. 2003.

[28] F. Ben Said, T. M. Hamdani, and A. M. Alimi, "Negotiation guiding framework: a recommendation for negotiation's designer model," in the 5th International Symposium on Computational Intelligence and Intelligent Informatics ISCIII, Floriana, Malta, 15-17 September 2011, pp. 105-110.

[29] G. Lai, K. Sycara and C. Li, "A decentralized model for automated multi-attribute negotiations with incomplete information and general utility functions," *Journal of Multi Agent and Grid Systems*, vol. 4, no. 1, pp. 45-65, Jan. 2008.

[30] F. Lopes, M. Wooldridge, and A. Q. Novais, "Negotiation among autonomous computational agents: principles, analysis and challenges," Artificial Intelligence Review, vol. 29, no. 1, pp. 1-44, Mar. 2008.

[31] S. Kraus, "Negotiation and cooperation in multi-agent environments," Artificial Intelligence, vol. 94, no. 1-2, pp. 79-97, July 1997.

[32] C. Yu, F. Ren, and M. Zhang, "An Adaptive Bilateral Negotiation Model Based on Bayesian Learning," in Complex Automated Negotiations:

Theories, Models, and Software Competitions, Studies in Computational Intelligence, Ito, Zhang, Robu, Matsuo (Eds.), Springer, 2013, vol. 435, pp. 75-93.

[33] C. Xueguang and S. Haigang, "Further Extensions of FIPA Contract Net Protocol: Threshold plus DoA," in Proceedings of the 2004 ACM Symposium on Applied Computing (SAC 2004), Nicosia, Cyprus, 14-17 March 2004, pp. 45-51.

[34] L. K. Soh and X. Li, "Adaptive, confidence-based multiagent negotiation strategy," in the 3rd International Joint Conference on Autonomous Agents and MultiAgent Systems (AAMAS'04), New York, USA, 19-23 July 2004, pp. 1048-1055.

[35] J. Harsanyi and R. Selten, "A generalised nash solution for two-person bargaining games with incomplete information," Management Sciences, vol. 18, no. 5, pp. 80–106, Jan. 1972.

[36] A. Quteishat, C. P. Lim, J. Tweedale, and L. C. Jain, "A Multi-Agent Classifier System Based on the Trust-Negotiation-Communication Model," in Applications of Soft Computing Part III, 2009, vol. 52 of the series Advances in Soft Computing, pp. 97-106.

[37] T. M. Hamdani, M. A. Khabou, and A. M. Alimi, "Conflict Negotiation Process with Stress Parameters Control for New Classifier Decision Fusion Scheme," in Proceedings of the 2010 International Conference on Image Processing, Computer Vision, & Pattern Recognition (IPCV 2010), Las Vegas, Nevada, USA, 12-15 July 2010, vol. 2, pp. 784-790.

[38] S. C. H. Hoi, J. Wang, and P. Zhao, "Libol: A library for online learning algorithms," Journal of Machine Learning Research (JMLR), vol. 15, no. 1, pp. 495-499, Feb. 2014.

[39] F. Ben Said and A. M. Alimi, "A New Online Feature Selection Method for Decision Making Problems with Ultra-High Dimension and Massive Training Data," *Journal of Information Assurance and Security (JIAS)*, vol. 11, no. 5, pp. 293-301, 2016.
[Online]. Available: http://LIBOL.stevenhoi.org

[40] A. Asuncion, and D. Newman. (2007). UCI Machine Learning Repository, [http://www.ics.uci.edu/~mlearn/MLRepository.html]. Irvine, CA: University of California, Department of Information and Computer Science.

[41] F. Ben Said and A. M. Alimi, "Multi Agent-Learner based Online Feature Selection System," IEEE International Conference on Systems, Man, and Cybernetics SMC'2016, October 9-12, 2016, Budapest, pp. 3652-3657.

[42] F. Ben Said and A. M. Alimi, "ANOFS: Automated Negotiation based Online Feature Selection Method", 15th International conference on Intelligent Systems Design and Applications (ISDA) (Class C), December 14-16, 2015, Marrakesh, Morocco, pp. 225-230.

[43] A. Das and M. Mahfuzul Islam, "SecuredTrust: A Dynamic Trust Computation Model for Secured Communication in Multiagent Systems," IEEE Trans. on Dependable and Secure Computing, vol. 9, no. 2, pp. 261-274, Mar.-Apr. 2012.

[44] F. Ben Said and A. M. Alimi, "OFSF-BC: Online Feature Selection Framework for Binary Classification," International Conference on Computer Systems and Applications, AICCSA'2016, November 29-December 02, 2016, Agadir, Morocco.

[45] P. Zhao, J. Wang, P. Wu, R. Jin, and S. C. H. Hoi, "Fast Bounded Online Gradient Descent Algorithms for Scalable Kernel-based Online Learning," *Proc. 29th International Conference on Machine Learning (ICML'12)*, pp. 2012.

[46] J. Wang, P. Zhao, and S. C. Hoi, "Exact Soft Confidence Weighted Learning," *Proc. 29th International Conference on Machine Learning (ICML'12)*, pp. 121-128, June/July 2012.

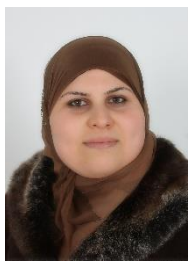

**Fatma Ben Said** was born in Sfax, Tunisia, in 1981. She received the Four-year University Degree in Applied Informatics in Management from Faculty of Economic and Management (FSEG), University of Sfax, Tunisia, in June 2004 and the master degree on Computer Science, Dedicated Computer System New Technologies (NTSID) from the National School of Engineers (ENIS), University of Sfax, Tunisia in July 2006. She received the Ph.D. graduate degree in Computing System Engineering, Research Group on Intelligent Machines, ENIS, University of Sfax, in 2017. She is an IEEE member since 2006.

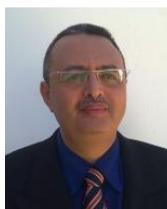

**Adel M. Alimi** was born in Sfax, Tunisia, in 1966. He graduated in Electrical Engineering 1990, obtained a Ph.D. and then an HDR both in Electrical & Computer Engineering in 1995 and 2000 respectively. He is now professor in Electrical & Computer Engineering at the University of Sfax.

His research interest includes applications of intelligent methods (neural networks, fuzzy logic, evolutionary algorithms) to pattern recognition, robotic systems, vision systems, and industrial processes. He focuses his research on intelligent pattern recognition, learning, analysis and intelligent control of large scale complex systems. He is associate editor and member of the editorial board of many international scientific journals (e.g. "IEEE Trans. Fuzzy Systems", "NeuroComputing", "Neural Processing Letters", "International Journal of Image and Graphics", "Neural Computing and Applications", "International Journal of Robotics and Automation", "International Journal of Systems Science", etc.). He was guest editor of several special issues of international journals (e.g. Fuzzy Sets & Systems, Soft Computing, Journal of Decision Systems, Integrated Computer Aided Engineering, Systems Analysis Modelling and Simulations). He is the Founder and Chair of many IEEE Chapter in Tunisia section, he is IEEE Sfax Subsection Chair (2011), IEEE ENIS Student Branch Counselor (2011), IEEE Systems, Man, and Cybernetics Society Tunisia Chapter Chair (2011), IEEE Computer Society Tunisia Chapter Chair (2011), he is also Expert evaluator for the European Agency for Research. He was the general chairman of the International Conference on Machine Intelligence ACIDCA-ICMI'2005 & 2000. He is an IEEE senior member.